\pdfoutput=1

\documentclass[11pt]{article}

\usepackage[preprint]{acl}

\usepackage{times}
\usepackage{latexsym}
\usepackage{times}
\usepackage{latexsym}
\usepackage{amsmath}
\usepackage{pgfplots}
\usepackage{hyperref}
\pgfplotsset{compat=1.18}
\usepackage{wrapfig}
\usepackage[T1]{fontenc}

\usepackage[utf8]{inputenc}

\usepackage{microtype}

\usepackage{inconsolata}

\usepackage{graphicx}

\title{Enhancing Knowledge Distillation for LLMs with Response-Priming Prompting}

\author{
  \textbf{Vijay Goyal\textsuperscript{1}},
  \textbf{Mustafa Khan\textsuperscript{1}},
  \textbf{Aprameya Tirupati\textsuperscript{1}},
  \textbf{Harveer Saini\textsuperscript{1}},
  \textbf{Michael Lam\textsuperscript{1}},
  \textbf{Kevin Zhu\textsuperscript{1}}}

  \author{Vijay Goyal \hspace{1cm} Mustafa Khan \hspace{1cm} Aprameya Tirupati \hspace{1cm} Harveer Saini \\
{\bf Michael Lam} \hspace{1cm} {\bf Kevin Zhu} \hspace{1cm}  \\
        Algoverse AI Research \\
        \texttt{kevin@algoverse.us}}

\begin{document}
\maketitle
\begin{abstract}
Large language models (LLMs) have demonstrated remarkable performance across a wide range of natural language processing (NLP) tasks. However, these models are often difficult to deploy due to significant computational requirements and resource constraints. Knowledge distillation (KD) is an effective technique for transferring the performance of larger LLMs to smaller models. Traditional KD methods primarily focus on the direct output of the teacher model, with little emphasis on the role of prompting during knowledge transfer. In this paper, we propose a set of novel response-priming prompting strategies applied in the knowledge distillation pipeline to enhance the performance of student models. Our approach fine-tunes a smaller Llama 3.1 8B Instruct model by distilling knowledge from a quantized Llama 3.1 405B Instruct teacher model. We apply LoRA optimization and evaluate on the GSM8K benchmark. Experimental results demonstrate that integrating reasoning-eliciting prompting into the proposed KD pipeline significantly improves student model performance, offering an efficient way to deploy powerful models in resource-constrained environments. We find that Ground Truth prompting results in a 55\% performance increase on GSM8K for a distilled Llama 3.1 8B Instruct compared to the same model distilled without prompting. A thorough investigation into the self-attention layers of the student models indicates that the more successful prompted models tend to exhibit certain positive behaviors inside their attention heads which can be tied to their increased accuracy. Our implementation can be found at \url{https://github.com/alonso130r/knowledge-distillation}
\end{abstract}

\section{Introduction}


Large language models (LLMs) have become widely used in various applications due to their high proficiency and adaptability toward performing diverse tasks. Recently, the parameter counts of LLMs have increased significantly, with models with well over 100 billion parameters becoming increasingly common \citep{minaee2024largelanguagemodelssurvey}. However, these models are resource intensive, which can render them inaccessible in settings where large-scale computation may not be available. This presents a challenge, as both LLM training and inference are computationally expensive, calling for methods to reduce the number of parameters without compromising performance \citep{chavan2024fasterlighterllmssurvey}. 



One such approach is knowledge distillation (KD), the process of training a smaller "student" model on the outputs of a larger "teacher" model to replicate the performance of the larger model in specific natural language processing tasks \citep{gu2024minillmknowledgedistillationlarge}. The output of the larger teacher model is first recorded and paired with the corresponding model inputs to form the \textit{transfer set}, a teacher-derived student model training dataset.  The student model is then fine-tuned on the knowledge transfer set. KD can thus potentially reduce the computational cost associated with running a larger model by partially transferring its behavior to a smaller, more accessible model.

In traditional knowledge distillation, the teacher model is directly prompted with the benchmark to create outputs for the transfer set. There has been a lack of current research on how specific prompting strategies during the formation of this transfer set affect the downstream performance of the student model. 

In this paper, we implement a set of novel response-priming prompting strategies in the knowledge distillation pipeline to study their efficacy of conferring performance benefits to the student model. We accomplish this by applying several engineered prompts to a teacher model to generate an elaborate set of logits representing the teacher model's solutions and thought processes for the questions. The student model is then fine-tuned on this set in order to learn the reasoning behind its teacher's answers. The student model is finally evaluated on a benchmark (evaluation split of the training dataset) to assess the effectiveness of this process. A quantitative evaluation of the student model strongly indicates that integrating prompting strategies with KD boosts reasoning performance over the unprompted distilled model and the non-distilled model. We intend for this work to lead to further studies exploring the ways in which prompting can increase the efficacy with which student models can learn from their teachers.






\section{Related Work}

Much research has been conducted on the modification of knowledge distillation to improve student model performance. \citet{mcdonald-emami-2024-trace} constructed a pipeline where the teacher model is prompted to generate a step-by-step problem decomposition on which the student model is trained, teaching the student model how to re-frame a question in a way downstream models can more easily understand and solve. This fine-tuning effectively aided the student model later in downstream problem-solving evaluations. 

Considerable work has also been done regarding the various effects of prompting, from improving reasoning capabilities to reducing hallucination. Zero-shot or few-shot prompting methods can be applied to models without the need for additional training \citep{kojima2023largelanguagemodelszeroshot}. Chain-of-thought, self-consistency, and the newer chain-of-knowledge and chain-of-verification methods are just a few of the many state-of-the-art prompting techniques used to increase performance, reduce hallucinations, and improve human interpretability \citep{chen2023universalselfconsistencylargelanguage, kojima2023largelanguagemodelszeroshot, li2024chainofknowledgegroundinglargelanguage, dhuliawala2023chainofverificationreduceshallucinationlarge}. 

In one study, chain-of-thought prompting was implemented into a KD pipeline to increase performance \citep{magister2023teachingsmalllanguagemodels}. However, this pipeline is outdated as it uses the cross entropy for soft loss, which is too coarse-grained for use with KD in LLMs  \citep{gu2024minillmknowledgedistillationlarge}. Although some success was achieved, our work expands on that by approaching the idea with new prompting strategies and more modern, LLM-specific loss.

\section{Method}

\begin{figure}[ht]
    \centering
    \includegraphics[width=0.5\textwidth]{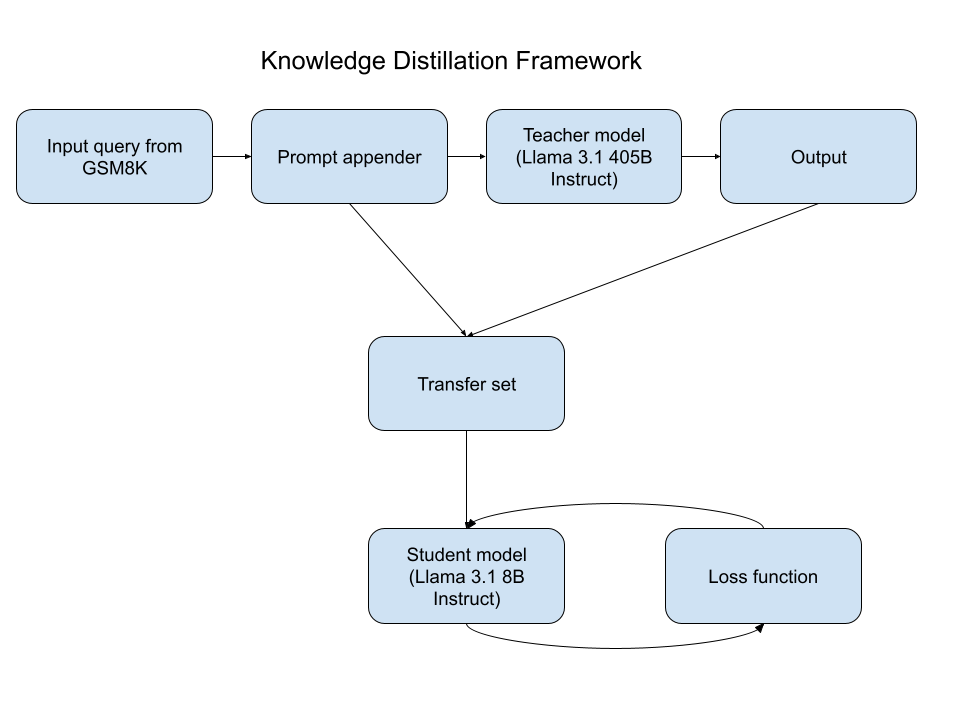}
    \caption{Visual representation of the knowledge distillation pipeline.} 
    \label{fig:kd-framework}  
\end{figure}

Our pipeline consists of two main stages. First, the teacher model responds to queries from the training dataset. Crucially, as opposed to in traditional knowledge distillation, we prompt the teacher model in this initial step to ensure that the student model has high-quality, reasoned data to learn from. Additionally, we use a large teacher model, since reasoning ability improves with higher parameter counts \citep{wei2023chainofthoughtpromptingelicitsreasoning}. We introduce three novel prompting strategies for this purpose, which are discussed in Section 3.1. Once the teacher model passes through a subset of the training data, the queries modified with prompts and the teacher model's responses are collected to form the transfer set. In the second stage, the student model is trained on the transfer set using the teacher model's logits and LoRA optimization \citep{hu2021loralowrankadaptationlarge}. The student model is fine-tuned for 2 epochs over a custom selection of GSM8K train split (the first 1319 questions) to produce the final model. We choose not to create training prompts for the student model because it is already fine-tuned on the teacher model to reason through problems, and over-complicating the training process could lead to decreased accuracy.  Figure \ref{fig:kd-framework} shows a summary of this method. 

\subsection{Prompting strategies}
\begin{figure*}
\centering
  \includegraphics[scale=0.2]{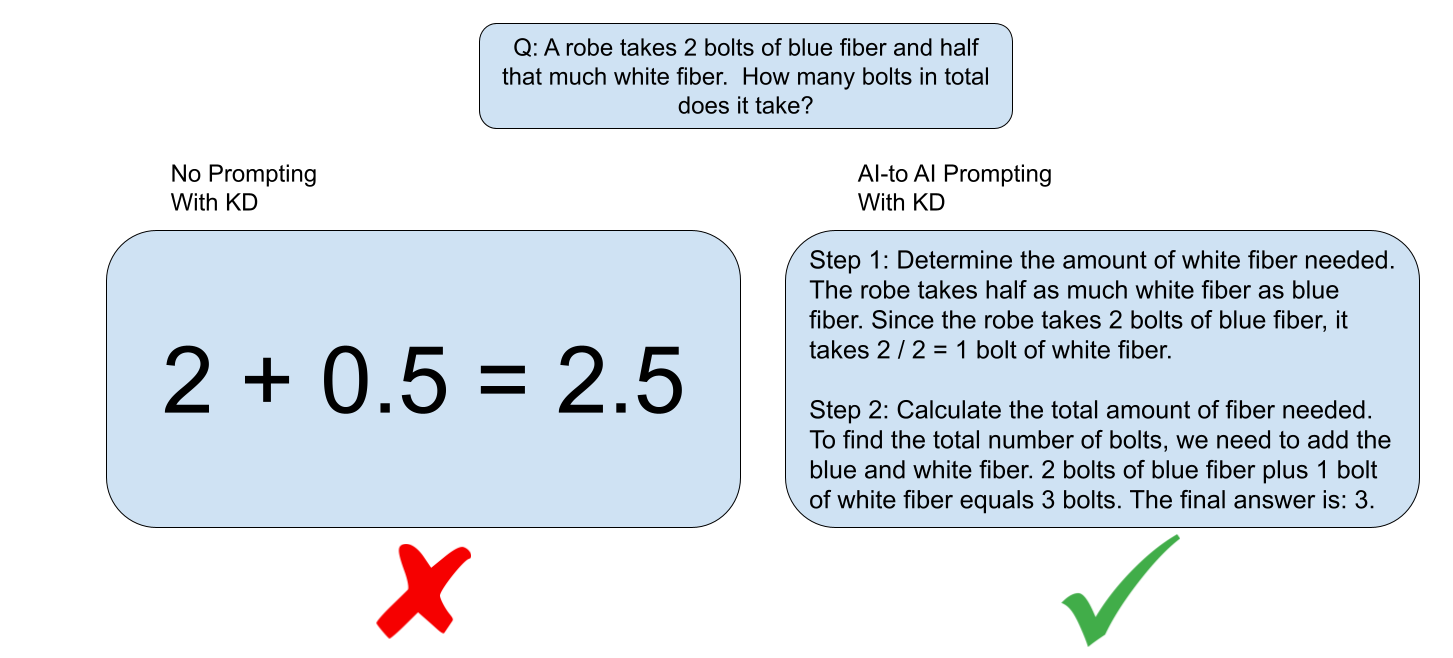}
  \caption{Example of the differences between prompted and unprompted model outputs.}
  \label{yes-no-prompt}
\end{figure*}
In this section, we introduce three novel prompting strategies to be implemented in the KD pipeline. Full prompts and more details can be found in the appendix. Figure \ref{yes-no-prompt} shows an example of the difference between unprompted and prompted ouputs. 
\paragraph{Teacher prompting} Intermediate reasoning steps have been shown consistently to support language models solving complex reasoning problems \citep{wei2023chainofthoughtpromptingelicitsreasoning}. Strategies such as  chain-of-thought and self-polish greatly increase performance by giving models crucial information through  these problem-solving steps \citep{chen2023universalselfconsistencylargelanguage, xi2024selfpolishenhancereasoninglarge}. We apply this idea by having the teacher model act as an actual teacher. This approach aims to cause the teacher model to explain its step-by-step process clearly, allowing the teacher's complex reasoning behavior to be more directly learned by the student. Then, the student should be able to work through problems more accurately, resulting in increased performance. 

\paragraph{Ground Truth prompting} Due to the growth of parameter counts in LLMs and an increased variety of available training data, many modern models understand being personified \citep{minaee2024largelanguagemodelssurvey}. We take advantage of this ability by telling the teacher model that it is a language model and that its outputs will be used to train smaller models. Through this approach, we intend for the teacher model to tailor its responses so that they can be easily learned from. 

\paragraph{Confidence prompting} This strategy is inspired by reverse KL divergence, which penalizes the student for underestimating probabilities where the target has a high probability mass (mode-seeking) \citep{tuananhleTuan}. This encourages the student to become more confident in the predictions it inherits from the teacher. In this strategy, we propose asking the teacher to check its final answer after generation, increasing confidence in the solution. This will lead to more pronounced spikes in the generated probability distributions, making the student more likely to be confident in correct predictions due to mode-seeking.

\section{Experimentation}
\subsection{Language Model Selection and Evaluation}
We employ Llama 3.1 405B Instruct, quantized using EXL2 to an average of 6 bits per weight, as our teacher model \citep{dubey2024llama3herdmodels}. We use six bit-per-weight quantization since the perplexity score of the quantized model remains equivalent to that of the 16 bit-per-weight full-sized model up until a context length of 100,000 tokens, which exceeds the output lengths for our purposes \citep{grimulkanllamaquant}. We use Llama 3.1 8B Instruct as a student model, with the BF16 weights provided in the default model \citep{dubey2024llama3herdmodels}. To isolate the effect of prompting strategies combined with KD, we evaluate all models using zero-shot prompting instead of the eight-shot prompting method used in the official evaluations of these models. Additionally, we assess all models using an output modifier inserted into the prompts to facilitate more efficient answer extraction. We fine-tune with LoRA optimization \citep{hu2021loralowrankadaptationlarge} with a matrix rank of 4 and a LoRA alpha value of 8, which targets the projection layers \((\text{k}_\text{proj}, \text{q}_\text{proj}, \text{v}_\text{proj}, \text{o}_\text{proj})\) of the self-attention mechanism. Finally, we select the first 2600 questions of GSM8K's train split as our training data and the full test split as our evaluation data \citep{cobbe2021trainingverifierssolvemath}.

\subsection{Training Loss Selection and Hyper-parameter Tuning}
We apply combined KD loss to the distilled models. The loss function and its hyper-parameters remain unchanged across models for consistency. We use cross-entropy loss for hard loss since it is computationally efficient and appropriate for the given task. For soft loss, we use reverse KL divergence for its mode-seeking behavior. This encourages the model to produce high-confidence predictions, a task with which models with fewer parameters typically struggle \citep{he2024lawnexttokenpredictionlarge}. 

\begin{flushleft}
\begin{minipage}{0.5\textwidth} 
    \begin{equation}
      \label{eq:eq1}
      p=\frac{z_{i}^{\text{teacher}}}{T}-\log(\sum_{j=1}^n \exp (\frac{z_{i}^{\text{teacher}}}{T}))
    \end{equation}
    \begin{equation}
      \label{eq:eq2}
      q=\dfrac{\exp (\dfrac{z_{i}}{T})}{\sum_{j=1}^n \exp (\dfrac{z_{i}}{T})}
    \end{equation}
    \begin{equation}
      \label{eq:eq3}
      \text{Loss}_\text{soft} = \sum_iq_i\log(\dfrac{q_i}{p_i})
    \end{equation}
    \begin{equation}
      \label{eq:eq4}
      \text{Loss}=\alpha \cdot \text{Loss}_\text{hard} + (1-\alpha) \cdot \text{Loss}_\text{soft} \cdot T^2
    \end{equation}
    \captionof{figure}{Equations representing the combination of soft and hard loss in knowledge distillation, starting from the teacher and student logits.}\label{fig:loss-func}
\end{minipage}
\end{flushleft}

To determine the $\alpha$ and temperature values for the loss function (Figure \ref{fig:loss-func}), we conduct a 50-iteration trial using Bayesian optimization. We initialize the student model in its default state. Then, we train on our custom selection of GSM8K’s train split for 2 epochs and evaluate on the full test split. After the trial, we select the set of hyper-parameters that produces the lowest validation loss. 

A lower $\alpha$ value indicates that the teacher model’s logits exert a positive influence on the student model, and vice-versa. Similarly, a high temperature value indicates the student model’s inability to understand the raw teacher probabilities. This suggests the need for softening to prevent the student from misinterpreting extreme probability values. 

Since the student model is a small fraction (1.98\%) of the scale of the teacher model, the temperature value of 5.9 falls near the midpoint of the range from 1.0 to 11.0. Our $\alpha$ value is 0.61, indicating that as defined by $\alpha$, the teacher logits had a strong positive effect on the performance of the student model.

\section{Results}
All trained student models were evaluated using 0-shot prompting, along with the exclusion of output modifiers used for ease of answer extraction.

\subsection{Accuracy Evaluation}

Quantitative evaluation of the student model strongly indicates that prompting strategies combined with KD increase reasoning performance. As shown in figure \ref{fig:eval-results}, models fine-tuned with Teacher and Ground Truth prompting show a significant increase in the amount of correct answers generated. Specifically, Ground Truth prompting shows a 55\% increase in accuracy over the unprompted distilled model, and a nearly 400\% increase over the non-distilled model. 
\scalebox{0.9}{
\begin{tikzpicture}

\begin{axis} [ybar,
    bar width = 16pt,
    ymin = 0, 
    ymax = 40, 
    enlarge y limits = {value = .25, upper},
    ylabel = {Performance on GSM8K (\%)},
    symbolic x coords = {No KD, No KD finetuned, Base KD, Confidence KD, Teacher KD, Ground Truth KD},
    x tick label style = {font = \small, text width = 1.7cm, align = center, rotate = 40, anchor = north east},
    xtick=data
    ]
    \addplot coordinates{
    (No KD,12.20) 
    (No KD finetuned, 25.01)
    (Base KD,30.62) 
    (Confidence KD,34.04) 
    (Teacher KD,42.30) 
    (Ground Truth KD, 48.14)};
\end{axis}
\end{tikzpicture}}

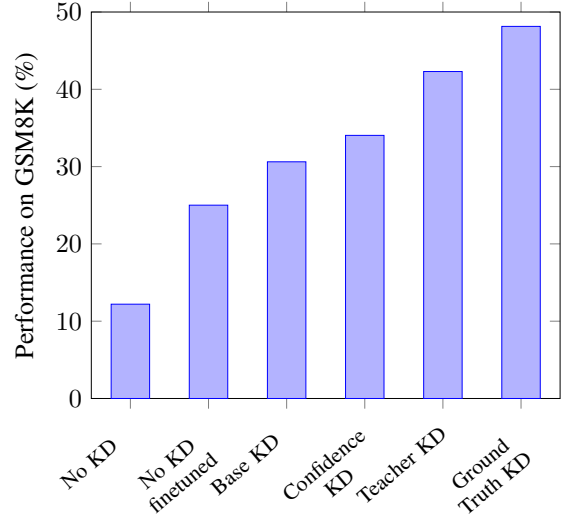
\captionof{figure}{Evaluation performance of all models based on answer accuracy.}\label{fig:eval-results}

\begin{figure*}
    \centering
    \includegraphics[width=\textwidth]{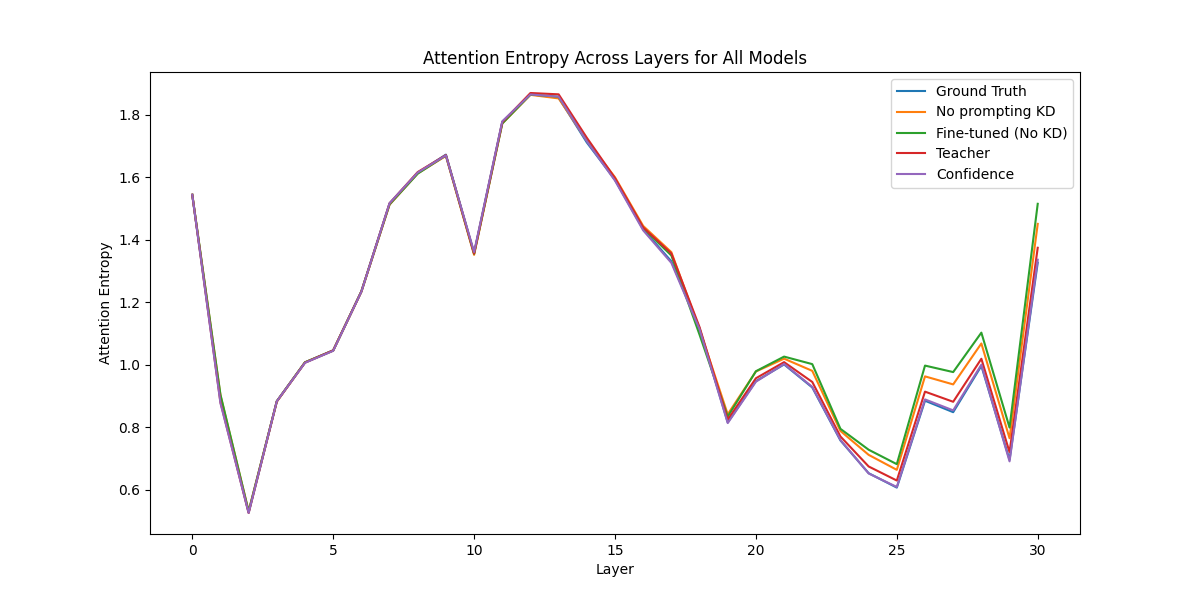}
    \caption{Visual comparison between the average entropy across all self-attention heads in a layer of the student/evaluated models.} 
    \label{fig:entropy}  
\end{figure*}

\subsection{Qualitative Analysis}
The quality of reasoning generated by the different types of models can vary quite drastically. In general, the rule is the higher the accuracy the more coherent and complete the logical reasoning the model creates is. 

The base model and fine-tuned (no KD) model have a tendency to not generate any intermediary reasoning steps and skip directly to the final answer, which is often incorrect. This is obviously not ideal for an LLM as they thrive off context, and this type of answer generates none.

The Base KD model is an improvement over the 2 previous models. However, it will occasionally produce the no reasoning, final answer only outputs that plague the base and fine-tuned model. It will occasionally output very high quality reasoning akin to that of the Teacher KD and Ground Truth KD, but not frequently enough for a large accuracy boost.

The Confidence KD model exhibits interesting reasoning skills. Although the steps it takes are often correct, the model has a high tendency to sidetrack and perform steps that are completely unnecessary, something that can negatively impact its final answer accuracy. The steps are also not very concise, leading to the model being cut off by the new token maximum quite often.

The Ground Truth KD and Teacher KD models appear to be the golden standard for reasoning, with concise, on task steps being generated. Crucially, these steps do not contain many logical reasoning fallacies, allowing the models to progress to a correct final answer effectively.

Coupled with the differences in final answer accuracy, the stark differences in reasoning capability across models leads to the need for a deeper dive into the root causes of these differences, particularly in the self-attention layers.

\begin{figure*}
    \centering
    \includegraphics[width=\textwidth]{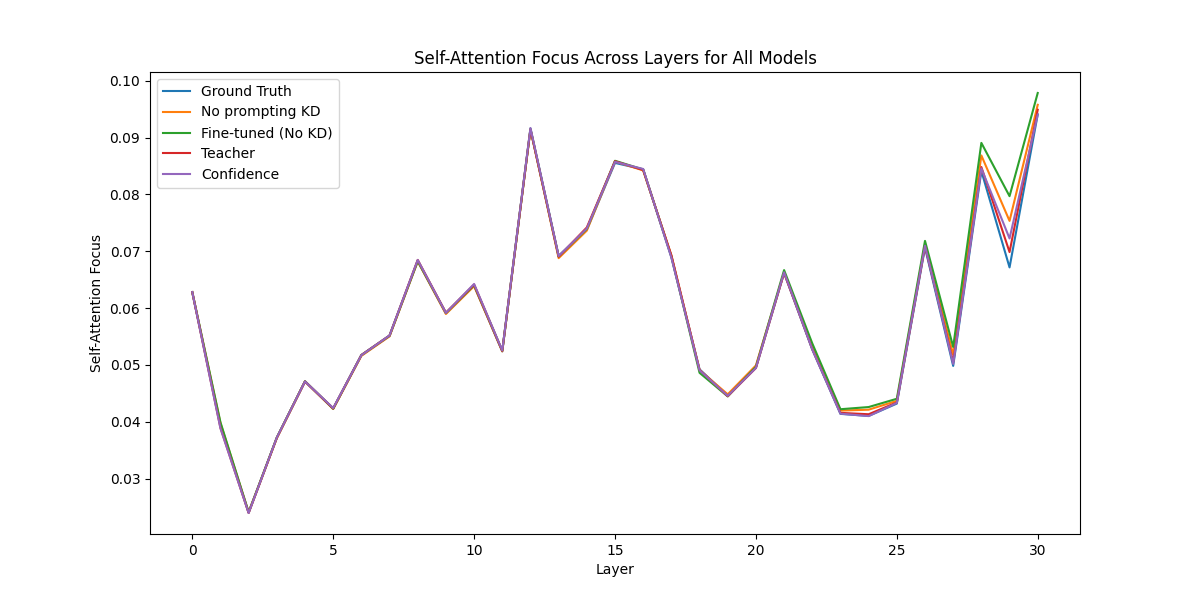}
    \caption{Visual comparison between the average self-attention (how much a token focuses on itself) per head per layer of the student/evaluated models.} 
    \label{fig:self_attention}  
\end{figure*}

\section{Investigation and Interpretation}

The model analysis results naturally lead to further questions about why specific prompts have such a powerful positive impact. As it is not clear why certain models are better at reasoning and arithmetic simply from looking at their responses and prompts, a more thorough investigation into the models is needed. The prompt used to calculate the attention scores used in the analysis is listed in the appendix.

\subsection{Attention Entropy}
\begin{flushleft}
\begin{minipage}{0.5\textwidth}
    \begin{equation}
      \label{eq:attention_probs}
      \text{Attention Probabilities}_{i,j} = \frac{\text{Attention}_{i,j}}{\sum_{k} \text{Attention}_{i,k}}
    \end{equation}
    \begin{equation}
    \small
      \label{eq:entropy}
      H_i = -\sum_{j} \text{Attention Probabilities}_{i,j} \cdot \log(\text{Attention Probabilities}_{i,j})
    \end{equation}
    \begin{equation}
      \label{eq:avg_entropy}
      \text{Entropy}_{average} = \frac{1}{N} \sum_{i=1}^{N} H_i
    \end{equation}
    \captionof{figure}{Equations for entropy calculation in attention distributions, starting from attention probabilities to average entropy across tokens.}\label{fig:entropy1}
\end{minipage}
\end{flushleft}

Figure \ref{fig:entropy1} defines the attention entropy across layers. Entropy is generally used as a measure of randomness or chaos in a system, and the demonstrated use is similar. By setting the attention score distribution of a self-attention head as the system, entropy can be used to determine how structured a model's self-attention is. A high or low score for this metric can be considered positive; however, the positioning of said score matters most. High entropy near the end of the layer sequence might suggest that a model has yet to finish interpreting the input tokens, leading to an incomplete understanding and/or missing context. The scores that are being compared to assess the models is the layer-by-layer score, where the average score per layer of self-attention heads is calculated. Looking towards the entropy comparison plot across all fine-tuned models (Figure \ref{fig:entropy}), it is clear that there exists a score divergence between different models that can be used as a method of quantifying the differences in output quality. Scores are relatively identical until the models approach the 20th layer. At that point, the scores diverge in the inverse of the final accuracy, with the Ground Truth prompted model ending with the lowest score. From this, it can be concluded that the Ground Truth prompted model is particularly good at finalizing its interpretation of the input sequence, which can have a powerful influence on the quality of reasoning. The rest of the tested models have scores corresponding to their accuracy ranking, except for Confidence prompting. Interestingly, the Confidence prompted model is not far behind the Ground Truth prompted model, which is not representative of its performance on the accuracy benchmark.

\subsection{Self-Attention Focus}
\begin{flushleft}
\begin{minipage}{0.5\textwidth}
    \begin{equation}
      \label{eq:self_attention}
      \text{Self-Attention}_i = \text{Attention}_{i,i}
    \end{equation}
    \begin{equation}
      \label{eq:self_attention_probs}
      \text{Self-Attention Probabilities}_i = \frac{\text{Self-Attention}_i}{\sum_{k} \text{Attention}_{i,k}}
    \end{equation}
    \begin{equation}
    \small
      \label{eq:avg_self_attention}
      \text{Self-Attention}_{average} = \frac{1}{N} \sum_{i=1}^{N} \text{Self-Attention Probabilities}_i
    \end{equation}
    \captionof{figure}{Equations for calculating self-attention focus, focusing on diagonal elements of the attention matrix.}\label{fig:self_attention1}
\end{minipage}
\end{flushleft}

Figure \ref{fig:self_attention1} defines Self-Attention Focus, a metric quantifying how much a token in an attention head focuses on itself. When all the tokens in a self-attention head primarily focus on themselves, a heat-map for this head will have a diagonal line pattern running through it. This metric calculates how close to said diagonal the attention weight distribution is. The average score for each layer is calculated, and this layer-by-layer score is used to compare the models against each other. A higher score indicates that the self-attention heads in this layer are generally having tokens focus on themselves, which is a neutral result. A lower score indicates that the tokens are looking elsewhere, indicating more contextual integration and a positive result. Figure \ref{fig:self_attention} clearly shows the differences between prompted and non-prompted models. From layers 23-25, all 3 prompted models dip lower then their non-prompted counterparts, indicating a more effective use of these layers to gain contextual understanding. After layer 26 the differences between models becomes more pronounced, with the more accurate models having a lower score in the last 2 layers. From this it can be concluded that models with a higher final answer accuracy (such as the Ground Truth prompted model) are better at using their final self-attention layers for contextual understanding. 

\begin{figure*}
    \centering
    \includegraphics[width=\textwidth]{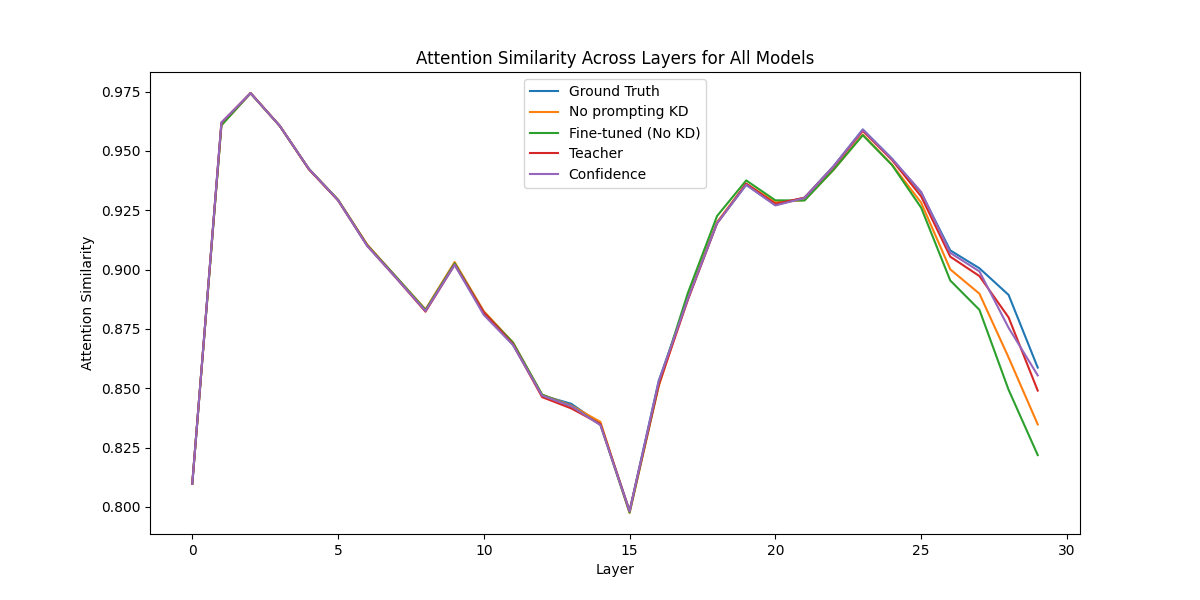}
    \caption{Visual comparison of the cosine similarity of an attention head in consecutive layers of the student/evaluated models.} 
    \label{fig:similarity}  
\end{figure*}

\subsection{Attention Distribution Similarity}
\begin{flushleft}
\begin{minipage}{0.5\textwidth}
    \begin{equation}
      \label{eq:similarity}
      \text{Similarity} = \cos(\theta) = \frac{\text{Flat\_norm}_l \cdot \text{Flat\_norm}_{l+1}}{\|\text{Flat\_norm}_l\| \|\text{Flat\_norm}_{l+1}\|}
    \end{equation}
    \begin{equation}
      \label{eq:avg_similarity}
      \text{Similarity}_{average} = \frac{1}{H} \sum_{h=1}^{H} \text{Similarity}_h
    \end{equation}
    \captionof{figure}{Equations for measuring attention distribution similarity across layers using cosine similarity of flattened and normalized attention matrices.}\label{fig:similarity1}
\end{minipage}
\end{flushleft}
Figure \ref{fig:similarity1} defines Attention Distribution Similarity, a metric used to calculate the inter-layer change in attention scores in the forward pass of a model. By calculating the cosine similarity of a self-attention head and its equivalent in the next layer, it is possible to determine how the attention scores of a model are changing layer-by-layer. A high score indicates that the self-attention score distributions are remaining constant, while a low score means that the distributions are changing. However, this metric relies on the layer's positional data to determine whether a high or low score is positive or negative. A low score near the beginning of the layer sequence can indicate that the model is not wasting layers to refine its interpretation of the sequence; while a low score near the end of the layer sequence could indicate that the model is not yet done refining its interpretation. In the plot visualizing the Attention Distribution Similarity scores for fine-tuned models (Figure \ref{fig:similarity}), all models have similar scores until layer 23, where key differences begin to show. The fine-tuned and non-prompted KD model end with low scores, potentially indicating unfinished interpretation of the sequence. The most accurate model, Ground Truth prompted KD, has the highest score, indicating that it is more confident in its sequence interpretation. Interestingly, Confidence prompted KD is again an outlier, with a score minorly above the Teacher prompted KD model.

\section{Discussion}
In addition to numerical performance gains on GSM8K evaluations, the student model also exhibits qualitative gains. We find that the student model exhibits certain behaviors during generation that are generally only found in models with significantly higher parameter counts, such as improved reasoning and contextual understanding \citep{naveed2024comprehensiveoverviewlargelanguage}. Interestingly, the quality of reasoning is not proportional to accuracy. Prompted distilled models create similar steps yet differ in accuracy. We propose 3 reasons for the increased accuracy of Ground Truth prompting. First, the average entropy of the self-attention score distribution is lower in the latter layers compared to its counterparts, indicating a more focused understanding of the input sequence. Second, the average Self Attention Focus of the model is lower in the final self-attention layers, indicating tokens are not focused on themselves, a sign of increased contextual integration. Thirdly, the high Attention Similarity score in the latter layers indicates that the model has finalized its interpretation of the input sequence by then, and will not be left with a sub-optimal understanding.




\section{Conclusion}
This study investigates the effect of response-priming prompting strategies applied to the teacher model in knowledge distillation. Our pipeline, which (1) prompts the teacher model to provide better learning material for the student and (2) fine-tunes the student on the teacher model's responses, introduces novel prompting techniques for use in KD. We find that prompting the teacher model for use during KD substantially improves accuracy on GSM8K by inducing specific behaviours in the self-attention layers of the student model which positively impacts mathematical skills along with reasoning skills.

\section{Limitations}
\paragraph{\textbf{Task and Domain Specificity}} Our experiments are conducted exclusively on the GSM8K benchmark, which focuses on mathematical problem-solving tasks. This narrow evaluation raises questions about the generalizability of our approach to other NLP tasks and domains. Further, our prompts are only tested on one pair of teacher and student models. Future research should explore the effectiveness of our prompting strategies across a broader spectrum of tasks and models to validate their universal applicability. Additionally, the $\alpha$ and temperature value hyper-parameter selection process is quite hidden and compute/time heavy. We believe that both hyper-parameters would need to be found using trial-and-error methods, starting from estimates found from the hyper-parameter count ratio of the student and teacher models.

\paragraph{Risks} There is an inherent risk with using LLMs in that they may generate inaccurate information. Further, there is a risk of the teacher model generating problematic responses, but we cannot evaluate this since we only access the teacher logits and do not continue to the final text outputs of the model. 


\bibliography{custom}

\appendix
\section{Prompts}
In this section, we list the exact prompts we use in experimentation: 

\paragraph{Teacher prompting} "As a teacher, guide your student through solving the question below. Provide a clear, simple explanation for someone unfamiliar with the problem." 
\paragraph{Ground Truth prompting} "As a smart language model, provide a detailed answer that is clear and feasible for smaller models to understand and learn from without losing too much detail. "
\paragraph{Confidence prompting} "Briefly double-check that your answer is correct once done." Note that this prompt is appended to the end of the question.
\paragraph{Attention analysis prompt}
"Janet’s ducks lay 16 eggs per day. She eats three for breakfast every morning and bakes muffins for her friends every day with four. She sells the remainder at the farmers' market daily for \$2 per fresh duck egg. How much in dollars does she make every day at the farmers' market?"
\section{Results table}
Refer to table \ref{tab:your_table_label}.
\begin{table}[h!]
  \centering
  \begin{tabular}{|c|c|}
    \hline
     Model Name& Performance on GSM8K \\ \hline
    Base & 12.20\% \\ \hline
    Finetuned base & 25.01\% \\ \hline
    Base KD & 30.62\% \\ \hline
    Confidence KD & 34.04\% \\ \hline
    Teacher KD & 42.30\% \\ \hline
    Ground Truth KD & 48.14\% \\ \hline
  \end{tabular}
  \caption{Exact results of all evaluations of finished models on GSM8K.}
  \label{tab:your_table_label}
\end{table}
\section{Example responses}
In this section, we detail example responses from a selection of models to a query from GSM8K. We provide analysis of the responses and discuss the differences between the outputs. 
\paragraph{Question} Janet’s ducks lay 16 eggs per day. She eats three for breakfast every morning and bakes muffins for her friends every day with four. She sells the remainder at the farmers' market daily for \$2 per fresh duck egg. How much in dollars does she make every day at the farmers' market?
\paragraph{No distillation response}
The answer is: \$22.00.
\paragraph{Base KD response}

Step 1: Calculate the total number of eggs laid by the ducks per day.
The ducks lay 16 eggs per day. Step 2: Determine the total number of eggs Janet eats and bakes
Janet eats 3 eggs for breakfast and bakes 4 eggs for muffins for her friends every day. The total number of eggs she consumes is 3 + 4 = 7. Step 3: Calculate the number of eggs Janet sells.
To determine how many eggs Janet sells at the farmers' market, subtract the number of eggs she consumes from the total number of eggs the ducks lay. eggs\_sold = 16 - 7 = 9. Step 4: Determine how much Janet makes at the farmers' market.
Janet sells her eggs for \$2 each. She sells 9 eggs. The amount Janet makes is calculated by multiplying the number of eggs sold by the price of an egg. money\_made = 9 * 2 = 18.
The final answer is: 18.
\paragraph{Teacher prompting KD response}
Step 1: First, calculate the total number of eggs laid by the ducks.
Janet's ducks lay 16 eggs per day.
Step 2: Calculate the number of eggs eaten by Janet for breakfast.
Janet eats 3 eggs every morning.
Step 3: Calculate the number of eggs used for baking muffins.
Janet bakes 4 muffins every day.
Step 4: Calculate the number of eggs left over for selling.
Total eggs laid by ducks = 16.
Eggs eaten for breakfast = 3.
Eggs used for baking muffins = 4.
Remaining eggs for sale = (16 - 3 - 4) = 9.
Step 5: Calculate the total money Janet makes by selling the eggs.
Janet sells the eggs at \$2 per egg.
Total money made by Janet per day = Number of eggs left * Price of each egg = 9 * 2.
Step 6: Calculate the value.
9 * 2 = 18.
\paragraph{Ground Truth prompting KD response}
Step 1: Calculate the total number of eggs laid by the ducks.
Janet’s ducks lay 16 eggs per day.
Step 2: Calculate the total number of eggs eaten and used for baking.
Janet eats 3 eggs for breakfast and bakes muffins using 4 eggs daily.
Step 3: Calculate the number of eggs left for sale after eating and baking.
16 eggs laid - (3 + 4) eggs used = 16 - 7 = 9 eggs.
Step 4: Calculate the total amount of money made by selling eggs.
She sells 9 eggs daily for \$2 each.
Step 5: Multiply the number of eggs sold by the selling price to find the total amount earned.
9 * \$2 = \$18.
\paragraph{Analysis}
The performance improvement from KD and from prompting is apparent in these responses. The non-distilled model fails to answer the question correctly and does not appear to think through the problem at all. The base KD model successfully responds to the question and shows step-by-step reasoning. This is likely a result of learning this behavior from the larger teacher model through distillation. Both the Ground Truth KD model and Teacher KD model also answer the question correctly and shows reasoning steps. However, we note that while both the prompted and unprompted KD models respond to the question correctly, the prompted models do so more concisely. Between the prompted models, Teacher prompting tends to over-explain compared to Ground Truth prompting. We believe that this plays a factor in better performance for Ground Truth prompting, as Teacher prompting may cause the model to confuse itself by over-reasoning. 

\section{Compute resources}
Teacher model: 4 NVIDIA H100 96GB for 2.5 hours per transfer set for a  total of 10 compute hours.

Student model training: 1 NVIDIA RTX 6000 Ada for 1 hour per model for a  total of 4 compute hours.

Student model inference: 2 NVIDIA RTX 6000 Ada for 1 hour per model for a  total of 4 compute hours.

Student model self-attention analysis: 1 NVIDIA RTX 6000 Ada for a  total of 1 compute hour.

\section{Licensing and Intended Use}
All work done with licensed scientific artifacts follows the terms of the licensing agreements and their intended use. Additionally, an effort was made to ensure that all artifacts used are open source in order to preserve repeatability of our experiments. The GSM8K dataset \citep{cobbe2021trainingverifierssolvemath} is available under the MIT license. Llama 3.1 is available under the Llama 3.1 Community Licensing Agreement \citep{dubey2024llama3herdmodels}. 

\section{Data Integrity}
We took steps to ensure our data was well-tested and was verified not to contain personally identifying information. Futher, we anonymized the GitHub repository provided.

\end{document}